\documentclass[conference]{IEEEtran}
\IEEEoverridecommandlockouts
\usepackage{cite}
\usepackage{amsmath,amssymb,amsfonts}
\usepackage{algorithmic}
\usepackage{graphicx}
\usepackage{textcomp}
\usepackage{xcolor}
\usepackage{hyperref}
\def\BibTeX{{\rm B\kern-.05em{\sc i\kern-.025em b}\kern-.08em
    T\kern-.1667em\lower.7ex\hbox{E}\kern-.125emX}}
\begin{document}

\title{AI Generated Text Detection}

\author{\IEEEauthorblockN{Adilkhan Alikhanov}
\IEEEauthorblockA{\textit{Department of Computer Science} \\
\textit{Nazarbayev University}\\
Astana, Kazakhstan \\
adilkhan.alikhanov@nu.edu.kz}
\and
\IEEEauthorblockN{Aidar Amangeldi}
\IEEEauthorblockA{\textit{Department of Computer Science} \\
\textit{Nazarbayev University}\\
Astana, Kazakhstan \\
aidar.aamangeldi@nu.edu.kz}
\and
\IEEEauthorblockN{Diar Demeubay}
\IEEEauthorblockA{\textit{Department of Computer Science} \\
\textit{Nazarbayev University}\\
Astana, Kazakhstan \\
diar.demeubay@nu.edu.kz}
\and
\IEEEauthorblockN{Dilnaz Akhmetzhan}
\IEEEauthorblockA{\textit{Department of Computer Science} \\
\textit{Nazarbayev University}\\
Astana, Kazakhstan \\
dilnaz.akhmetzhan@nu.edu.kz}
\vspace{1em} \\  
\IEEEauthorblockN{Nurbek Moldakhmetov}
\IEEEauthorblockA{\textit{Department of Computer Science} \\
\textit{Nazarbayev University}\\
Astana, Kazakhstan \\
nurbek.moldakhmetov@nu.edu.kz}
\and
\IEEEauthorblockN{Omar Polat}
\IEEEauthorblockA{\textit{Department of Computer Science} \\
\textit{Nazarbayev University}\\
Astana, Kazakhstan \\
omar.polat@nu.edu.kz}
\and
\IEEEauthorblockN{Galymzhan Zharas}
\IEEEauthorblockA{\textit{Department of Computer Science} \\
\textit{Nazarbayev University}\\
Astana, Kazakhstan \\
galymzhan.zharas@nu.edu.kz}
}

\maketitle

\begin{abstract}
The rapid development of large language models has led to the increased amount of AI-generated text and students using the LLM generated texts as their work, which violates academic integrity. This paper presents an evaluation of AI-text detection including both standard machine learning models and transformer-based architectures. We utilize 2 datasets, HC3 and DAIGT v2, and then build a unified benchmark and apply a topic-based data split to prevent information leakage. This ensures robust generalization across unseen domains. Our experiments show that tf-idf logistic regression achieves reasonable baseline accuracy of 82.87\%. However, deep learning models actually outperform it. BiLSTM classifier achieves the accuracy of 88.86\%, and DistilBERT achieved a similar accuracy of 88.11\% with the highest ROC--AUC score of 0.96, demonstrating the highest performance. The results show that contextual semantic modeling is far superior to  lexical features and that it is important to mitigate topic memorization using evaluation protocols. Our limitations are mainly related to dataset diversity and computational constraints. In the future, we plan to work on expanding dataset diversity and then utilize parameter efficient fine tuning methods such as LoRA. We can also explore smaller or distilled models and utilize more efficient batching and hardware-aware optimization. 
\end{abstract}


\section{Introduction}
\subsection{Motivation}
The emergence of Large language models (LLMs) has led to an unprecedented spread of AI-produced text in academic, professional, and online settings. The distinction between human and machine generated text has increasingly become a challenge, especially with modern detectors often being unstable, having biases, or easily being compromised. Continually improving the detection technologies is essential to maintain the academic integrity and transparency, as well as reducing the abuse of generative AI systems. This creates a strong motivation to create robust, data-driven approaches that can be scaled across different domains and model types.
\subsection{Problem Statement}
The goal of this work is to design and evaluate a unified system for classifying whether a given text sample is AI-generated or human-written. For this purpose, we considered multiple modelling approaches—including classical machine learning, sequence-based neural networks, and transformer-based architectures and used two datasets (HC3 and DAIGT v2). The study aims to assess model performance, analyze generalization under topic-grouped splits, and identify limitations of existing detection pipelines.
\section{Related Work}
Due to the rapid development of LLMs, a lot of research has been done in this area. The authors have tried a variety of methods, which we can broadly categorize into the following categories: traditional ML and statistics-based methods and Deep learning based methods. 

\subsection{Traditional ML and statistical methods}
In the beginning of research authors have heavily relied on extracting linguistic features from texts and then feeding them into Machine learning models. The features generally fall into lexical, structural, and complexity categories. 
One of the extracted lexical features is POS tags such as NOUN, VERB, ADJ, and so on. In total, 18 POS tags were extracted \cite{yadagiri2024linguistic,yadagiri2024deep}. It was found that AI generated texts usually have more number of NOUN, VERB, ADP, AUX tags as compared to human generated texts  \cite{yadagiri2024deep}. Another lexical features are the voice usage, active vs passive voices, and text style. It was found that AI generated texts tend to have more linking words \cite{yadagiri2024linguistic}. The extracted statistical features include Readability metrics, lexical analysis, predictive metrics.  The readability metrics such as Flesch Reading Ease score and Gunning Fog Index evaluate how complex a text is \cite{yadagiri2024linguistic}. The lexical analysis includes extracting the following features: average line length, vocabulary size, and word density \cite{yadagiri2024linguistic}. As for the predictive metrics, Perplexity and Burstiness were computed for texts \cite{yadagiri2024linguistic}. 
After extracting features Prova \cite{islam2023nlp} trained and test  such models as XGBoost and SVM and found that they can achieve a reasonable performance. For example, SVM and XGBoost had accuracies of  81\% and  84\% respectively. However, they are not good enough to trust such systems.

\subsection{Deep learning based methods}
Deep learning models were primarily based on the transformer architecture and they demonstrated much superior performance. 

Yadagiri et al. (2024) trained RoBERTa on HC3 dataset by fusing linguistic and statistical features with the word embeddings \cite{yadagiri2024linguistic}. They extracted the linguistic and statsitical features from the texts such as Average Line Length, Word Density, Part-of-Speech (POS) tags, Flesch Reading Ease score, Gunning Fog Index, Perplexity and so on \cite{yadagiri2024linguistic}. Then, they concatenated these features with the word embeddings generated by the RoBERTa. This methodology helped them achieve the accuracy of 99\%. They injected a small trainable low rank matrices to the model's attention layers and fine-tune the model on the HC3 dataset training only the weights in the matrix\cite{yang2024lora}. This helped them achieve the accuracy of 91\% while cutting the training time and other costs by a lot. Roand LoRA Optimization: To reduce training time and computational overhead, one study employed Low-Rank Adaptation (LoRA) to fine-tune the RoBERTa-base model on the HC3 dataset \cite{yang2024lora}. LoRA freezes the pre-trained weights of RoBERTa and injects small, low-rank trainable adapters between specific layers \cite{yang2024lora}. This strategy allowed for efficient fine-tuning while maintaining high performance \cite{yang2024lora}.

Another paper used BERT model and fine-tuned it on their own private dataset for binary classification task \cite{wang2023bert}. As a result, they achieved the accuracy of 97\%. Another paper that also used BERT managed to achieve the accuracy of 93\% on their own private data \cite{islam2023nlp}. The difference between them is most likely explained via the difference in data quality and number. Another author proposed a hybrid method involving Bi-LSTM and Attention \cite{blake2025bilstm}. Authors first extracted POS tags from texts and then passed through an embedding layer, which learned the grammar and other patterns. Then, they are passed through convolutional layers to extract local patterns and then are passed through Bi-LSTM layer to capture long-range dependencies. On top of that, attention layer was applied which determined which POS tags mattered more. At the end, they managed to hit the accuracy of 88\% \cite{blake2025bilstm}. 
\section{Datasets}

This project relies on two public datasets often used in
AI text detection studies : HC3 and DAIGT v2. These include matched samples of human writing vs. machine-made text from various areas. This way, we can test how well the tools work on different subjects or tones.

\subsection{HC3 Dataset}

HC3 (Human-ChatGPT Comparison Corpus) is a high-quality benchmark designed to compare human versus machine responses under fixed conditions using side-by-side analysis. The dataset consists of questions and answers gathered from online websites like Reddit’s ELI5, plus Quora or StackExchange. With every query, one response comes from a person and another is made by ChatGPT.

This results in a balanced dataset containing approximately 37k human replies and 37k ChatGPT-generated outputs. The labeling is fixed and unambiguous, ensuring high label consistency. HC3 also includes topic information: the \texttt{source} field allows grouping into domains such as \texttt{HC3\_reddit\_eli5}, \texttt{HC3\_medicine}, and \texttt{HC3\_finance}.

From a language perspective, HC3 gives brief answers and then
mainly informative or clarifying in tone. The limited instruction-reply
layout helps HC3 work better with traditional ML techniques.
Lexical trends in human versus ChatGPT output often show strong similarity
results. A key drawback is that HC3 uses just one LLM - meaning no variety in model output (ChatGPT). Possibly causing skewed outcomes due to narrow training focus or reduced adaptability across different tasks
content produced by recent large language models. 

\subsection{DAIGT v2 Dataset}

DAIGT v2 is a broad collection made public for an AI-written text identification task. Around 44,800 brief writings - crafted by people or various language models - are included, such as versions from the GPT series, openly available transformer networks, and proprietary artificial intelligence tools. Instead of manual tagging, identifiers come from contest-related data details, which ensures consistent marking between human and machine-produced entries.

The dataset shows uneven labels - 61\% human, 39\% AI - mirroring actual usage better than evenly split sets like HC3. Every entry includes a \texttt{source} tag indicating the subject area (for example, \texttt{DAIGT\_Driverless cars}, \texttt{DAIGT\_Exploring Venus}, or \texttt{DAIGT\_Extracurricular activities}), enabling clustering by theme.

\begin{table}[ht]
\centering
\caption{Comparison of the HC3 and DAIGT v2 Datasets}
\begin{tabular}{lcc}
\hline
\textbf{Property} & \textbf{HC3} & \textbf{DAIGT v2} \\
\hline
Total samples & $\sim$74k & 44.8k \\
AI model sources & ChatGPT only & Multiple LLMs \\
Human/AI ratio & 50/50 & 61/39 \\
Text style & Q\&A & Essays \\
Topic diversity & Medium & High \\
Label quality & High & High \\
\hline
\end{tabular}
\label{tab:datasets}
\end{table}

\subsection{Topic-Based Splitting}

Rather than applying a random split, which could cause information leakage 
between similar topics, we adopt a topic-based partitioning strategy. Entire topics are assigned to the training, validation, or test sets, ensuring that:

\begin{itemize}
    \item the test set contains \emph{unseen} domains,
    \item models cannot rely on memorization of topic-specific cues,
    \item we effectively simulate out-of-distribution (OOD) generalization.
\end{itemize}

The final split proportions look like this:
\begin{itemize}
    \item \textbf{69.2\%} of training samples,
    \item \textbf{20.1\%} of validation samples,
    \item \textbf{10.7\%} of test samples.
\end{itemize}

This method of splitting actually provides a more challenging evaluation setup compared to traditional random sampling. Additionally, it highlights differences between classical and transformer based approaches.

\section{Methodology}

\subsection{Dataset Preparation}

A single dataset was formed by combining HC3 \cite{hc3dataset} and DAIGTv2 \cite{daigtv2dataset}. This gave 124,195 text samples under 20 different topic categories. HC3 subset provided about 74,000 samples in the form of Q\&A domains such as Reddit ELI5, finance, medicine, and general knowledge. DAIGTv2 consists of  44,800 short essay samples, which cover topics such as distance learning, driverless cars, and community service. Each of the samples of the texts underwent data preprocessing stage to eliminate null entries and proper string formatting.

Initial attempts of data preprocessing with shuffling and random splitting resulted in the model learning only topic specific vocabulary and producing too optimistic accuracies even in Logistic Regression. The model was not learning to detect the AI writing style, and thus we faced the data leakage problem. In order to avoid data leakage instead of randomly sampling we used a topic-based splitting strategy. Each topic category was put in either training, validation, or test partitions. Five major sources (HC3\_reddit\_eli5, HC3\_finance, DAIGT\_v2\_Distance learning, DAIGT\_v2\_Seeking multiple opinions, and HC3\_open\_qa) were used as the training set (85,897 samples, 69.2 percent). Validation set consisted of eight topics (24,987 samples in total, 20.1 percent) and the test set consisted of seven new topics with 13,311 samples (10.7 percent). This divisiveness made sure that models could not take advantage of topic specific pattern of vocabularies and be trained on more general features of stylistic markers between human and AI generated texts.

\subsection{Model Architectures}

We evaluated three model paradigms representing increasing levels of architectural complexity: classical machine learning, recurrent neural networks, and transformer-based models.

\subsubsection{Logistic Regression Baseline}

The baseline model employed TF-IDF vectorization with unigrams and bigrams (ngram\_range=(1,2)) and English stop word removal, followed by logistic regression classification. Hyperparameter optimization via 5-fold cross-validation with GridSearchCV explored vocabulary sizes of \{15000, 25000, 35000\}, regularization strengths C $\in$ \{0.1, 1, 10\}, and penalty types \{L1, L2\}. The loss function uses was Binary Cross-Entropy.

\subsubsection{Bidirectional LSTM}

The BiLSTM architecture \cite{hochreiter1997lstm} consisted of an embedding layer (dimension=128, vocabulary size=30,000), a bidirectional LSTM layer processing sequences up to 600 tokens, dropout regularization, and two fully connected layers with ReLU activation leading to sigmoid output. The vocabulary size of 30,000 covered 97.65\% of word occurrences, and the sequence length of 600 tokens encompassed 95\% of all texts. Training used binary cross-entropy loss with Adam optimizer over 15 epochs with early stopping (patience=3). Grid search over 36 configurations explored LSTM units $\in$ \{64, 128, 256\}, dropout rates $\in$ \{0.2, 0.3, 0.5\}, batch sizes $\in$ \{128, 256\}, and learning rates $\in$ \{0.0005, 0.001\}. 

\subsubsection{DistilBERT}

We fine-tuned DistilBERT-base-uncased \cite{sanh2019distilbert}, a 6-layer transformer with 66 million parameters pre-trained on English Wikipedia and BookCorpus. Input sequences were tokenized to a maximum length of 512 tokens, covering over 90\% of samples without truncation. The model employed AdamW optimizer with linear warmup scheduling and gradient clipping (max\_norm=1.0). Hyperparameter search explored learning rates $\in$ \{2e-5, 3e-5\}, batch sizes $\in$ \{8, 16\}, with fixed training duration of 3 epochs. 

\subsection{Experimental Setup}

We conducted all experiments on a PC with NVIDIA RTX 5080 GPU with 16GB of VRAM. Random seeds were set at 42 in all experiments for reproducibility purposes. We selected models based based on validation accuracy, and the best checkpoint was saved for final evaluation on the test set.

\section{Results}

This part shows the experimental results of each of the three modeling methods, performance measures, training dynamics and error behavior.

\subsection{Overall Performance Comparison}

Table~\ref{tab:results} is a summary of all model performances on the test set. The baseline of the logistic regression obtained 82.87 percent accuracy which shows that lexical features in their own alone give enough information for classification. The BiLSTM added value to this baseline achieving 88.86 percent accuracy and a ROC-AUC of 0.94. DistilBERT had the highest ROC-AUC of 0.96 and the highest accuracy of 88.11, which suggests that it ranks better although it has slightly lower accuracy than BiLSTM.

\begin{table}[htbp]
\caption{Model Performance Comparison on Test Set}
\label{tab:results}
\centering
\begin{tabular}{lcccc}
\hline
\textbf{Model} & \textbf{Accuracy} & \textbf{ROC-AUC} & \textbf{Train (s)} & \textbf{Infer (s)} \\
\hline
Logistic Reg. & 82.87\% & -- & 201.04 & 0.01 \\
BiLSTM & 88.86\% & 0.94 & 4682.84 & 1.21 \\
DistilBERT & 88.11\% & 0.96 & 9554.14 & 64.67 \\
\hline
\end{tabular}
\end{table}

\subsection{Per-Class Performance Analysis}

The preciseness, recall and F1-scores of each model are provided in Table~\ref{tab:classification}. The balance of the class distribution in the test set was unequal, 9,717 human samples (73.0), and 3,594 AI-generated samples (27.0).

\begin{table}[htbp]
\caption{Per-Class Classification Metrics on Test Set}
\label{tab:classification}
\centering
\begin{tabular}{llccc}
\hline
\textbf{Model} & \textbf{Class} & \textbf{Precision} & \textbf{Recall} & \textbf{F1} \\
\hline
Logistic Reg. & Human & 0.93 & 0.83 & 0.88 \\
Logistic Reg. & AI & 0.64 & 0.82 & 0.72 \\
\hline
BiLSTM & Human & 0.93 & 0.91 & 0.92 \\
BiLSTM & AI & 0.78 & 0.82 & 0.80 \\
\hline
DistilBERT & Human & 0.97 & 0.87 & 0.91 \\
DistilBERT & AI & 0.72 & 0.92 & 0.81 \\
\hline
\end{tabular}
\end{table}

Logistic regression also had the minimal level of accuracy when it comes to AI recognition (0.64), which means a high number of false alarms of human-written text. BiLSTM was the most balanced model in both classes, and the highest accuracy in AI detection (0.78) and, therefore, the lowest number of false positives (839 out of 13,311 samples). DistilBERT showed the best recall when it comes to AI-generated text (0.92), successfully identifying 3,322 of 3,594 AI samples but with only 272 false negatives- the lowest of any model. This feature is what makes DistilBERT especially good at tasks od identifying AI text.

\subsection{Confusion Matrix Analysis}

\begin{figure}[htbp]
\centering
\includegraphics[width=0.32\columnwidth]{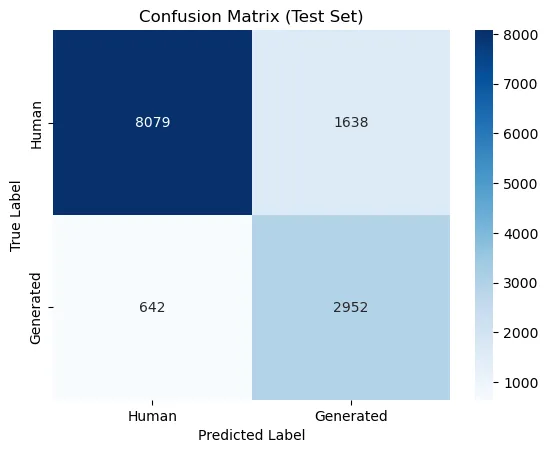}
\includegraphics[width=0.32\columnwidth]{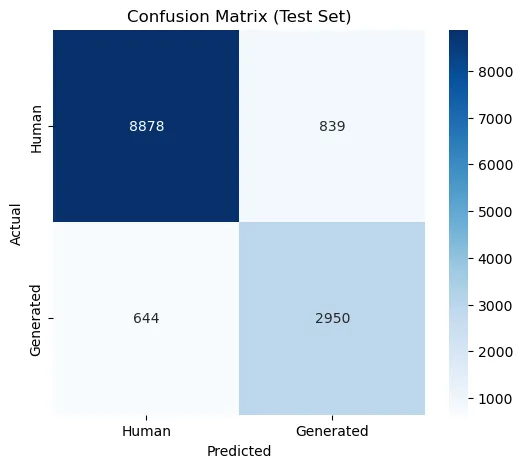}
\includegraphics[width=0.32\columnwidth]{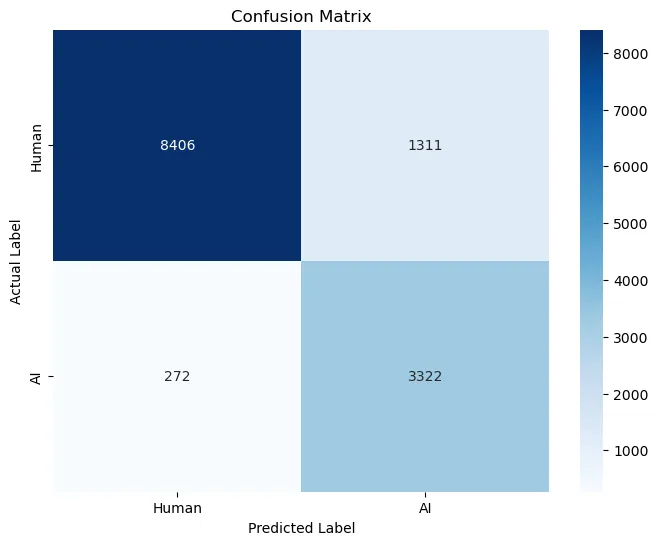}
\caption{Confusion matrices for (left) Logistic Regression, (center) BiLSTM, and (right) DistilBERT on the test set.}
\label{fig:confusion}
\end{figure}

 The confusion matrices of all three models have been displayed in Fig.~\ref{fig:confusion}. The produced logistic regression baseline had 1,638 false positives (human text that was incorrectly classified as AI), and 642 false negatives (AI text missed). BiLSTM minimized number of false positives to 839 at the expense of false negative (644). DistilBERT had the least number of false negatives (272) but had more false positives (1,311) as compared to BiLSTM.

\subsection{Training Dynamics}

There was a significant difference between types of training behavior. In the topic-based split, logistic regression had severe overfitting, with 99.22 percentage training accuracy and 82.87 percentage test accuracy, which corresponds to 16.35 percentage point of overfitting.

The training of BiLSTM exhibited typical instability in the validation metrics as both the validation loss and accuracy varied significantly within the epochs though the training loss was gradually declining. Recurrent architecture is normally associated with this behavior, but this was overcome by applying early stopping, which ended training once three consecutive epochs showed no improvement in validation accuracy.

DistilBERT showed the least volatile training dynamics, and the validation accuracy of the initial two epochs was around 95.3 percent, and after which it continued to achieve good consistency. The learned representations were pre-trained and thus it quickly converged with the model making a significant generalization after three epochs. But gradual loss of validation as training went on, indicated weak overfitting, but did not significantly impair ultimate performance.

\subsection{ROC Analysis}

\begin{figure}[htbp]
\centering
\includegraphics[width=0.48\columnwidth]{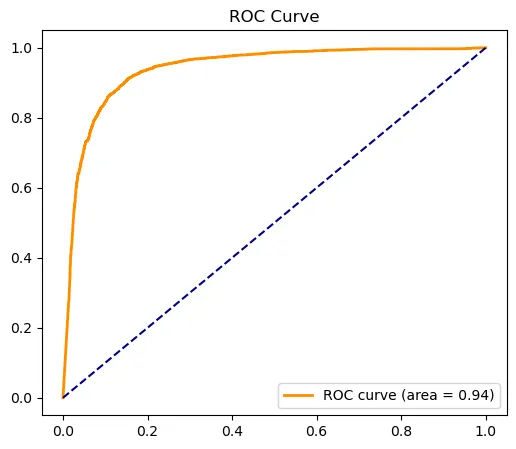}
\includegraphics[width=0.48\columnwidth]{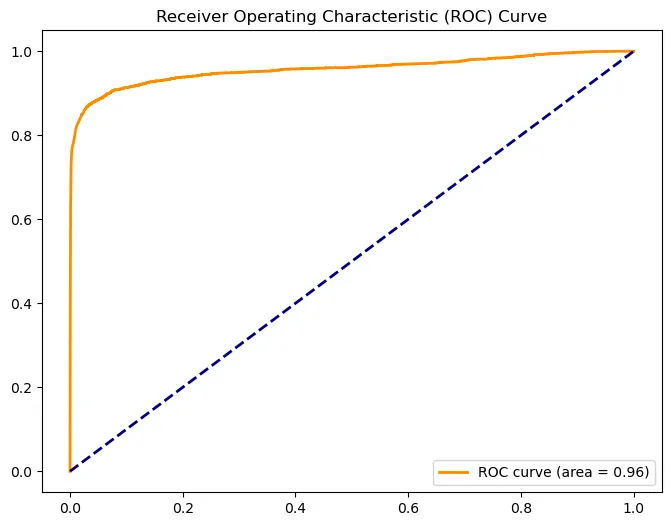}
\caption{ROC curves for BiLSTM (AUC=0.94) and DistilBERT (AUC=0.96).}
\label{fig:roc}
\end{figure}

The ROC curves of BiLSTM (AUC=0.94) and DistilBERT (AUC=0.96) are placed in Fig.~\ref{fig:roc}. The superior ranking property of DistilBERT is indicated by the larger AUC- in general, the model provides more confidence scores to AI-generated samples compared to BiLSTM even with the same classification threshold. The property is useful where one needs to have calibrated confidence estimates or varying decision thresholds.

\subsection{Computational Trade-offs}

The three methods had radically different computational issues. Training logistic regression required about 3.4 minutes with nearly immediate inference (0.01s on 13,311 samples) and is therefore useful in resource-constrained applications. BiLSTM took 78 minutes to train and 1.21 seconds to run inference, which is a fair compromise to gain better performance. DistilBERT required the largest computational cost of 159 minutes to train and 64.67 seconds to run inference--about 50 times slower than BiLSTM to run at inference time. These differences show the trade-offs involved in practice between model complexity and deployment limitations, especially with real-time users.

\section{Discussion}
\subsection{Model Performance}
Across all experiments, the three models---logistic regression (TF--IDF), DistilBERT, and BiLSTM---achieved high performance in detecting AI-generated vs.\ human-written text. The logistic regression baseline was a good one, which was confirming that simple lexical cues already provide meaningful signal. However, the deep learning models consistently outperformed it. The BiLSTM reached a ROC-AUC of $\sim$0.94, while DistilBERT achieved the best overall results with $\sim$0.96 ROC-AUC and $\sim$95--96\% validation accuracy. These improvements highlight the advantage of models which take contextual semantics (BiLSTM) and large-scale pre-training (DistilBERT).

\subsection{Generalization \& Model Behavior}
To make sure that the evaluation is fair, topic-grouped splitting was employed to ensure that the models are not trained to memorize topic-specific vocabulary. In this tougher arrangement, every model showed good performance, which indicated that it learned stylistic encoding and not topic artifacts. BiLSTM and DistilBERT seemed to be more sensitive to subtle patterns of sequencing or context, whereas logistic regression was more dependent on superficial patterns. The training dynamics of DistilBERT were also the most stable and quickly converged, with little overfitting, highlighting the advantage of transformer pre-training.

\subsection{Impact of Data Processing}
The combination of HC3 and DAIGT v2 produced a more heterogeneous dataset and minimized the bias of a single source of models. The sample cleaning and grouping prior to splitting prevented leakage and generated a more lifelike classification task. Despite these limitations, the logistic baseline was effective and the BiLSTM model and especially the transformer model made use of more intricate syntactic and semantic features. Early training and regularization provided some stability in training, particularly to the BiLSTM. On the whole, models based on transformers offered the best generalization of performance.

\subsection{Why DistilBERT Instead of TinyBERT}
We have chosen DistilBERT as it provides the best size-performance-training stability. The DistilBERT paper indicates that the model preserves the performance of BERT and is 40\% smaller and 60\% faster, respectively, which makes it a powerful compressed version with no significant loss of performance.

In comparison, TinyBERT offers an even higher level of compression, but usually requires task-specific distillation; i.e., the model needs to be re-distilled on the task of interest to reach competitive accuracy. In the absence of such an expensive teacher--student fine-tuning step, TinyBERT frequently performs worse than DistilBERT in classification tasks that involve subtle semantic distinctions, such as AI-generated versus human-generated text. Since we are interested in high out-of-the-box generalization and do not have substantial time to perform multi-stage distillation, DistilBERT was a more trustworthy option.

\section{Limitations \& Future Work}
\subsection{Weaknesses}
Our detectors perform very well, yet several limitations should be considered. 

\begin{enumerate}[label=\textbf{First}, leftmargin=*, align=left]
\item \textbf{Training data can cause model-specific bias:} The majority of generated AI samples are based on a small number of LLMs (mostly ChatGPT, along with a few from DAIGT v2). Consequently, the models cannot be easily generalized to unseen or future LLMs, particularly those that generate text with different styles or produce post-edited human--AI hybrid text. The human-written samples also represent a limited range of styles (primarily Q\&A format), meaning that unconventional or highly creative human writing may be misclassified.
\end{enumerate}

\begin{enumerate}[label=\textbf{Second}, leftmargin=*, align=left, start=2]
\item \textbf{Limited to English language:} Our study only considers English text, which does not account for multilingual or cross-domain scenarios. This limits real-world applicability where writing varies significantly across languages and genres. We also do not address error trade-offs in detail: false positives (classifying human text as AI-generated) could have serious implications, but we prioritized overall accuracy and AUC over careful decision-threshold calibration.
\end{enumerate}

\begin{enumerate}[label=\textbf{Third}, leftmargin=*, align=left, start=3]
\item \textbf{Computational constraints:} Our experimental design was shaped by limited computational resources. We were unable to incorporate very large datasets (e.g., LLM-Detect) or test substantially larger models. Hardware limits (16 GB VRAM) constrained sequence length and batch size, which may affect performance on long documents. Time and compute budgets also prevented extensive hyperparameter searches or the evaluation of full-scale BERT/GPT-based detectors. These factors may conceal certain weaknesses that would only emerge in larger-scale experiments.
\end{enumerate}
\subsection{Future Directions}
Future studies ought to increase the breadth and variety of datasets, such as the multilingual corpora or other diverse writing styles by humans and AI texts provided by newer or less popular LLMs. The evaluation and robustness would be improved with the introduction of adversarial examples i.e. paraphrased, mixed or deliberately obfuscated AI output. False positives also need to be addressed, e.g. decision-threshold tuning, confidence calibration or abstain/flag mechanisms might help alleviate high stakes misclassification.

On the modeling front, a larger transformer, domain-adapted architecture (e.g., RoBERTa), or GPT-style discriminators can help. Unless we have to resort to expensive computational cost, parameter-efficient tuning techniques such as LoRA may allow training large models. Another potentially successful trend is a combination of transformer embeddings and linguistic features (e.g., perplexity, grammar metrics). It is possible that the overall reliability can be more reliable through ensemble methods, as they use the complementary capabilities of logistic regression, BiLSTM, and transformers.

Lastly, in the context of detectors deployed into real-world environments, concept drift will have to be addressed when the LLMs are developing. A pipeline of active-learning that has new cases of AI-generated text added to it regularly would enable to retain the effectiveness of the model over the long term. In general, the future trends in the development of AI-text detection should be focused on the growth of datasets, their robustness, and the exploration of more sophisticated or hybrid schemes.

\section{Conclusion}
This paper presented a comparative analysis of AI-generated text detection in three modelling paradigms, including classical machine learning, recurrent neural networks and transformer-based paradigms. By combining HC3 and DAIGT v2 datasets and employing topic-grouping division to prevent information leakage, we developed a strong evaluation model that not only evaluates true generalization as opposed to memorizing topics but also evaluates generalization without memorizing a specific topic. 

Our experiments demonstrate that while simple TF-IDF with logistic regression achieves reasonable baseline performance (82.87\% accuracy), deep learning approaches substantially improve detection capabilities, with BiLSTM reaching 88.86\% accuracy and DistilBERT achieving the best overall results with 88.11\% accuracy and 0.96 ROC-AUC. These findings confirm that contextual and semantic modeling provides significant advantages over lexical features alone. We find that TF-IDF with logistic regression gives a good baseline (82.87\% accuracy), whereas deep learning models are significantly better. Specifically, the BiLSTM model achieves an accuracy of 88.86\%, and DistilBERT provides the best overall accuracy of 88.11\% and ROC-AUC of 0.96. These results indicate that contextual and semantic information models are much more effective compared to lexical feature-driven models.

However, the constraints that were found - model-specific bias, English-only scope, and computational constraints - point to valuable directions to be pursued in the future. Improving the dataset diversity, dealing with adversarial robustness, and studying more complex or hybrid networks are all important (though not the only) steps to implementing reliable AI text detection to actual practice where the impact of misclassification can be severe.

\section*{Code Availability}

The complete source code, model notebooks, trained checkpoints, and dataset used for the experiments in this study are publicly available in a Git repository. The repository can be accessed at: \url{https://github.com/crusnix/ai_text_detector_final}

\section*{Contributions}

The contributions of the authors are listed below, following the CRediT (Contributor Roles Taxonomy).

\begin{itemize}
    \item \textbf{Adilkhan Alikhanov:} Investigation, Conceptualization, Writing – Review \& Editing.
    
    \item \textbf{Aidar Amangeldi:} Formal Analysis, Visualization, Writing – Review \& Editing.
    
    \item \textbf{Diar Demeubay:} Methodology, Formal Analysis, Writing – Original Draft, Conceptualization.
    
    \item \textbf{Dilnaz Akhmetzhan:} Project Administration, Validation, Writing – Original Draft, Conceptualization.
    
    \item \textbf{Omar Polat:} Conceptualization, Visualization, Writing – Review \& Editing.
    
    \item \textbf{Galymzhan Zharas:} Formal Analysis, Visualization, Conceptualization, Writing – Review \& Editing.
    
    \item \textbf{Nurbek Moldakhmetov:} Methodology, Formal Analysis, Writing – Original Draft, Conceptualization.
\end{itemize}

\end{document}